**What do the metrics mean? A critical analysis of the use of Automated Evaluation Metrics in Interpreting**

*Jonathan Downie, Independent researcher jonathan.downie@gmail.com*

*Joss Moorkens, SALIS/ADAPT Centre, Dublin City University joss.moorkens@dcu.ie*

1. Introduction

With the growth of interpreting technologies, from remote interpreting and Computer-Aided Interpreting to automated speech translation and interpreting avatars, there is now a high demand for ways to quickly and efficiently measure the quality of any interpreting delivered. A range of approaches to fulfil the need for quick and efficient quality measurement have been proposed, each involving some measure of automation. This article will examine these recently-proposed quality measurement methods and will discuss their suitability for measuring the quality of authentic interpreting practice, whether delivered by humans or machines.

Two broad categories of method have emerged. The first category is semi-automated quality evaluation, which combines some form of human annotation and processing with computerised analysis. The second category borrows from established measures of the performance of machine translation (MT) systems to produce fully automated quality evaluation. This article will seek to answer two key research questions, based on available primary and secondary research:

1) How reliable are automated evaluation metrics for interpreting (hereafter 'interpreting AEMs'), given existing theoretical and practical knowledge of interpreting and specifically discussions of interpreting quality?

2) What role might interpreting AEMs play in moves towards fair, reliable, and efficient measures of interpreting quality, for both human interpreting and automated speech translation?

A fundamental assumption of this present article is that, since the aim of interpreting AEMs is to facilitate a quick measurement of interpreting quality irrespective of how it is delivered, it makes sense to treat the measurement of the quality of human interpreting and the measurement of the quality of automated speech translation systems as parallel, or even largely synonymous, endeavours. While it is true that the earliest AEMs can be found in translation and were not created with the aim of evaluating the usefulness of translation systems in real-world cases, the use of interpreting AEMs has tended not to differentiate between the comparison of different interpreting systems and the evaluation of how a system might perform in real-life.

One place a division *is* evident is in the different types of interpreting that have been evaluated. A clear difference can be seen in the measurement of the quality of largely unidirectional interpreting, as is often expected to be found in simultaneously interpreted events, compared to dialogic interpreting, which is most often found in events that involve a greater degree of interaction between the participants. Recent discussions have foregrounded the theoretical weakness of traditional interpreting settings – such as conference interpreting, community interpreting, and public service interpreting – as predictive categories (Mikkelson 1999; Gile and Napier 2020; Buzungu and Hansen 2020; Downie 2021). In this article, therefore, the interpreting mode involved will be used as the separating factor. However, given the above literature and explicit calls for findings from different parts of interpreting studies to be brought together (Gile and Napier 2020; Buzungu and Hansen 2020; Downie 2023), this article will end with an examination of some foundational practices for the use of interpreting AEMs, no matter the interpreting mode examined. Before any of this, however, it is important to explain what is meant by interpreting AEMs and the different methods available.

## 2. Introducing Automated Evaluation Metrics

There are two main approaches that have been taken towards creating interpreting AEMs. These in turn can be split into two further categories each, giving a total of four approaches. The first set of interpreting AEMs involve devising automated or semi-automated evaluations of interpreting *without* relying on metrics from MT. The impetus to create new automated evaluations that are specifically built for interpreting is justified by the view that interpreting involves different challenges to translation. Thus, Xiaojun Zhang argues that "it's unfair to evaluate human interpreting quality with MT metrics for interpreting itself is disfluent … Speech translation is normally an interactive process, and it is natural that it should be less than completely automatic" (2016: 2). The second set of interpreting AEMs reuse the methods and models used in the evaluation of MT. This second choice means equating interpreting and translation as linguistic practices. The two options will now be examined in order, taking non-MT metrics first.

### *2.1 Non-MT metrics: partial evaluations and full evaluations*

The primary difficulty posed when attempting to create AEMs without recourse to those designed for evaluating MT is to decide what should be measured. Two approaches have emerged. This first is to seek to measure a single variable. Thus, Zhang (2016) employed semi-automated evaluation of the closeness of semantic frames in the source and target text and Yu and van Heuven (2017; 2021) used acoustic indicators, such as articulation rate and silent pauses, to predict how interpreters would score on fluency.

These partial metrics therefore do not seek to provide an overall picture of interpreting quality but concentrate on a single factor, usually after having justified its importance in the literature. Thus, Zhang (2016: 2–3) appeals to linguistics and the view that the meaning of a word is found by examining the context in which it occurs. Yu and van Heuven similarly appeal to the known link between the perceived fluency of interpreting and its perceived quality (2017: 49–50) to justify the need for an objective measure of fluency against which human judgments can be compared. Where they differ from Zhang is that, while Zhang admits that the straight comparison between the semantic frames of the source and target text is a "naïve idea" (2016: 4), Yu and van Heuven conceive of automated fluency measures as a possible time and cost saving in interpreter assessment and a useful tool for trainee interpreters to use in their own development (2017: 50).

Single metric approaches therefore come with both a justification of the importance of a given metric and stated limitations of to what extent this single metric provides an indication of overall quality. It is only natural then that some scholars would opt for a more generalised approach, seeking to produce an automated or semi-automated evaluation of the overall quality of the interpreting provided.

Wang and Yuan (2023) and Wang and Wang (2024), for example, sought to build quality criteria models using machine learning to predict which interpreters would pass a screening examination. In both cases, the researchers began by setting the model parameters based on crowdsourced judgments of MT quality (Wang and Yuan 2023: 3) before allowing the model to develop. Their final results, in which their best algorithm could only predict human scores with 62.96% accuracy (ibid, pp.8-9), suggest that human scoring is more complex than imagined and that such models are not yet suitable for the pedagogical purposes intended. The difficulties inherent in building AEMs without using tools developed for the evaluation of MT may be the reason why some researchers have turned to MT AEMs. The use of these AEMs will now be examined, after a brief excursus on their history.

*2.2 Metrics from Machine Translation (MT AEMs)*

The first MT AEMs were developed by the US Department of Defence Research and Development group in order to avoid the problems of subjectivity, time, and cost associated with human evaluation (White et al. 1994). The objective of these AEMs was to quickly tell system developers whether their system had improved after a change (Koehn 2010). The most widely used metric, BLEU (Papineni et al. 2002), involves the comparison of an MT-proposed sentence with a human reference or 'gold standard' translated sentence. The closer the MT proposal comes to the reference translation, the higher the BLEU score will be. However, they are not compared as whole sentences, but by using n-grams. Machine translated sentences are broken into contiguous chunks of one, two, and three words, each of which is evaluated against the reference. The idea here is to allow for changes in word order. Then, there is a brevity penalty for very short sentences. The BLEU score of each sentence is averaged for a larger text to give a document-level score.

*2.3 The known issues with MT metrics.*

The problems with these early MT metrics became fairly clear almost as soon as they were used. The variability of human scores that prompted the development of AEMs exists because there is no single correct way to translate a sentence. In time, BLEU scores allowed multiple reference translations, but inevitably not all correct options can be covered. BLEU scores tend not to correlate well with human judgments (Callison-Burch et al., 2006; Tatsumi 2010). However, AEMs were only ever intended to assist system developers and not to give a true and objective evaluation of MT quality.

In time, 'shared tasks', competitive events that pit developers' MT systems against one another using the same training data, moved away from AEMs (Graham et al. 2017). The state of the art for AEMs moved from comparisons with reference translations (now derisively termed 'string-based metrics') to pre-trained neural systems (similar to those mentioned in Section 2.1), some with word embeddings that can identify collocates used in training data, the idea being that they can better generalise. Prominent MT researchers such as Kocmi et al. (2021) believe that string-based AEMs, and specifically BLEU, have incentivised developers to build systems that will produce high BLEU scores. "Do not use BLEU", they conclude, as "it is inferior to other metrics, and it has been overused" (ibid., p. 479).

Despite the growing scepticism regarding string-based AEMs and BLEU scores, they were used as the basis for Microsoft's infamous claim of reaching human parity for the quality of the news output of their Chinese-English MT system (Hassan et al. 2018). This claim prompted a strong response from the MT community, with Läubli et al. (2020) highlighting that the professional human translations used in the Microsoft experiment contained significantly fewer errors, as Hassan et al. (2018) themselves had noted in their paper, and that "the availability of linguistic context, and the creation of reference translations have a strong impact on perceived translation quality" (ibid, p. 669). Context-aware MT and evaluation have since become major aims for MT research and development.

The development of AEMs is a good example of how a view of translation – and interpreting – as a solely linguistic act "is common and persistent among NLP researchers" (Miyata et al. 2022: 2). The communicative context is not considered, but then the aim for AEMs was never to provide an authoritative score for translation, but rather to help developers to iteratively improve their systems. Taking a broader view of ecological validity that considers the repercussions of a score and its use as a basis for action (see Messick 1988 and Moorkens 2024), the use of automated metrics for MT is problematic, and many in the MT community now frown upon their use without parallel human evaluation. The likes of Mathur et al. (2020: 4893) argue against the use of AEMs "as the sole basis to draw important empirical conclusions". This makes it a little surprising when those criticisms are reduced to a footnote, such as in the work of Lu and Han (2023).

## 2.4 The Latent Quality Definition Used in AEMs

At this point, it is important to remember that, whichever kind of AEM is used, the same latent definition of quality is present. The act of trying to define an automated or semi-automated scheme for providing a repeatable, context-free definition of quality reflects a view of quality that is grounded in mathematical relationships between the source and target text. Thus, quality in AEM terms is always backwards-looking – asking whether there exists a sufficient relationship between the source and target text. The backwards-looking nature of AEMs is most noticeable when string-based AEMs, such as BLEU, are used since these are reliant on the presence of a reference translation. Thus, it is fair to say that such AEMs are not measuring how well a given translation might fit a specific purpose, but how closely it matches a previous translation.

The definition of quality in interpreting AEMs may therefore be expressed as the extent to which the target text reflects a given set of features in the source text. Exactly which features are measured, how they are measured, and the justification given for measuring these features varies from metric to metric but the underlying logic remains the same. Defining and measuring the source-target relationship remains the holy grail of AEMs, under the assumption that this relationship is the most important marker of quality.

This emphasis on the source-target relationship means that any idea of translation – and by extension, interpreting – being part of wider social purposes is not given any relevance. Thinking through how and for what purposes a translation or interpreting system will be used simply does not figure in the research methodology. The fact that a system scores highly on a specific test or that independent sentences are marked well by human judges does figure. Social aspects of interpreting are therefore implicitly excluded. This exclusion may well explain, for example, why human and machine evaluations of interpreting have tended to correlate more highly when the rubrics used for human evaluation are more highly structured (Lu and Han 2023; Wang and Fantinuoli 2024; Wein *et al.* 2024). It is less that some point of objectivity is being reached and more that the method of human evaluation is approaching that of the machines, precisely by progressively excluding social or one might say, subjective, factors.

Put another way, interpreting AEMs, like the translation AEMs on which some of them are based, trade ease of replication and seeming objectivity for the removal of any sense of social context from the act of evaluation. Indeed, some AEM designers have been clear that their aim was to provide a metric that has the benefits of speed and freedom from human variation (Han and Lu 2023: 1066; Lu and Han 2023: 110; Wang and Fantinuoli 2024: 2). Yet, since interpreting is an embodied human act (see Pöchhacker 2024), it would seem that human variation is an intrinsic part of both the delivery of interpreting and any subsequent evaluation. More specifically, the very variability that AEMs seek to minimise or eliminate reflects the context-bound nature of interpreting. It is ironic that, while there are ongoing methodological efforts to eliminate such variability and context-dependence via AEMs, discussions of quality in interpreting studies have become more aware of the foundational nature of such variability and of reckoning with context-dependence for providing useful accounts of the quality of authentic interpreting. It is to these discussions that this article will now turn.

### 3. Interpreting Quality

Any assessment of AEMs must be done within the framework of an overarching conversation about interpreting quality. It is important to understand what should be measured before attempting to measure it. Yet the voluminous literature on interpreting quality is yet to arrive at a consensus as to what and how to measure. In their recent handbook entry on interpreting quality, Macías and Zwischenberger point to two divergent views, the first concentrates on "the relation between participants in an abstract interpreting context using their attitudes, needs and points of view", while the other assesses the product of interpreting (Macías and Zwischenberger 2022: 243).

Largely, this shadows the classification of research on stakeholder expectations and responses by Downie (2015). In this latter case, research was classified along two dimensions: whether the research was tied to a specific authentic context or was about generic expectations of interpreting and whether clients were asked their expectations of interpreters or their response to a particular instance of interpreting (ibid, p. 24). Beyond Downie's (2015) classification and within the scheme by Macías and Zwischenberger (2022) is the research that has sought to provide a linguistic assessment of the correspondences of the source and target text in interpreting and the importance of any differences between the two. Such research ranges from the simple examination of errors in Mwinuka et al (2022) to the complicated measurement of audience responses to the manipulation of intonation and fidelity in the experimental work carried out by a group under Collados Aís (Collados Aís 1998; Collados Aís *et al.* 2007; Collados Aís 2016). Thus, product-based research on interpreting quality does not simply mean performing a semantic comparison but assessing what any specific patterns in the interpreted text might mean. For the most part, researchers within Interpreting

Studies are increasingly in agreement that quality in interpreting is a social variable, as much as it is a linguistic one, and thus that the interpreting context must play a role (Macías and Zwischenberger 2022: 252–253).

Given this increasingly social perspective and the now-accepted distinction between abstract impressions and context-specific evaluations, it is helpful to further sub-divide extant quality research according to the interpreting mode being used and the degree to which the interpreting examined is authentic. No distinction is made here between research carried out within or outside of Interpreting Studies, nor whether it is humans or machines being evaluated. Since the use of AEMs to measure quality has already been discussed, the research examined here will cover only work that has not used AEMs. Before exploring product-based evaluations of interpreting quality, it is useful to provide a brief overview of survey research on expectations of quality.

### 3.1 Survey research on interpreting quality

The idea of using surveys of interpreting users has not always been popular in Interpreting Studies. If it is taken as a truism that interpreting users need interpreting because they do not have the requisite language skills to manage without it, it stands to reason that they might not be the most attuned to what good interpreting is. In the words of Miriam Shlesinger, they may not know "what's good for them" (Shlesinger 1994: 126). Yet, as Downie argues in his survey of research on expectations of users and interpreters alike, the mere commercial logic that interpreters rely on satisfied clients for their income makes understanding client expectations important (Downie 2015: 21).

#### 3.1.1   Survey research in monologic interpreting

Survey research on monologic interpreting has sought to cover the views of both interpreters and clients, with methodological issues causing difficulties for direct comparability. The first such research is the now-famous study by Bühler (1986), which asked members of the AIIC Committee for the Admission and Language Classification of Applicants to rate the importance of a set of 16 items: native accent, fluency of delivery, logical cohesion of utterance, sense consistency with original message, completeness of interpretation, correct grammatical usage, use of correct terminology, use of appropriate style, pleasant voice, thorough preparation of conference documents, endurance, poise, pleasant appearance, reliability, ability to work in a team, positive feedback of delegates. A key assumption of this research was that the views of these key interpreters reflected wider patterns of expectations (ibid, p. 231). This assumption, which we might label as an assumption of homogenous quality, would be tested by later research.

Indeed, the work of Ingrid Kurz (1989; 1994; 2001), which used eight of Bühler's criteria (sense consistency with the original, correct use of terminology, logical cohesion, pleasant voice, grammatical correctness, native accent, fluency, and completeness of information) still represents the most thorough attempt at testing whether expectations are homogenous. While interpretations of these results differ somewhat (see the discussion in Chiaro and Nocella 2004; and Pochhacker 2005), what is clear is that they do not represent definitive proof that the assumption of homogenous quality was safe. Neither is such proof found in larger scale studies, such as the work by Moser (1995), which covered 85 conferences but did not include enough responses from any single conference to provide clear results as to any differences between expectations at different events (ibid, p. 5). Of course, absence of evidence is not evidence of absence. In fact, since the beginning of survey research on users of monologic interpreting, researchers have reported theoretical and methodological issues that may make it more difficult to arrive at a definitive answer as to whether expectations are truly homogenous (Mack and Cattaruzza 1995; Chiaro and Nocella 2005).

These theoretical and methodological difficulties arise from the known problem of creating well-defined quality criteria that are understood similarly by both respondents and researchers while being meaningful in a wide-range of contexts (Mack and Cattaruzza 1995: 47; Downie 2015: 27; Downie 2016: 42–43). The crux of the problem is that, without such well-defined and mutually well-understood criteria, it is difficult to provide a clear interpretation of results (Kurz 2001: 397, 403). The nature of this problem was laid bare in the work of Diriker, who found that, while all clients at a simultaneously interpreted philosophy conference said that the most important quality criteria was "fidelity to the meaning of the original speech", no two agreed what this meant (Diriker 2004: 131–147).

The problem came to a head in the work of Eraslan (2011), who found that the responses given by interpreting users were highly dependent on the nature of the questions posed. When users were asked questions about interpreting in general, they tended to give responses that painted interpreters as uninvolved conduits of information (ibid, p. 77-80). When users were asked how interpreters should respond to the challenges they faced when interpreting the specific event that the users were attending, they opted for a much more visible and involved role for interpreters, including adding explanations of culture-specific terms (ibid, p. 81) and clearing up misunderstandings (ibid, p. 83). In this present article, this contradiction between generic views of interpreting and responses to questions over what interpreters should do in specific situations will be labelled the Eraslan Effect, as it represents a theoretical and methodological barrier to the creation of a measurement of quality that includes social factors. On the basis of this effect, Eraslan argues that it is important to view expectations as reflections of the needs and requirements of individual contexts, rather than as criteria against which quality can be measured as an abstract concept (ibid, p. 200).

Put simply, surveys of client expectations of monologic interpreting provide a snapshot of the views at a specific event or time and do not necessarily offer any insight into the measurement of quality as a whole. On the contrary, the need to account for possible differences in how different survey items could be understood and the differences between general criteria and those specific to a given event suggest that quality itself may be a localised phenomenon. It may be more correct to refer to interpreting quality as perceived by clients at a specific event, rather than as adherence to a set of external criteria.

### 3.1.2   Survey research in dialogic interpreting

Much the same pattern can be found in survey research of dialogic interpreting contexts. Once again, such research has moved from attempts to use surveys to define and measure quality criteria to them being used to understand expectations in specific scenarios. This is evident both in surveys of users of dialogic interpreting and of the interpreters involved.

Typical in this regard is the survey by Lee (2009). This research sought the views of legal professionals and interpreters on a variety of matters, from the role of the interpreter to the quality of the interpreting they had experienced. It shows a slight reoccurrence of the Eraslan Effect, with most legal professionals viewing interpreters as translation machines (67%) or facilitators of communication (54%), while only 6% viewed them as cultural experts (ibid, p. 43). Yet, when asked if interpreters should convey the meaning of culture-bound expressions, 47% of legal professionals said they should, with 30% expecting terms of address to be treated similarly, 28% wishing interpreters to convey the meaning of gestures, 22% cultural customs and behaviours, and 19% cultural concepts.

This survey therefore offers a composite and partially contradictory picture of quality through the eyes of users, with interpreters not being required to be cultural experts yet possibly being expected to convey the meaning of cultural expressions. Interpreters were expected to be facilitators of communication yet were expected to reproduce grammatical errors, even though almost half of legal professions admitted that such errors made them think there was a problem with interpreting (ibid, p. 49).

Further surveys on expectations in legal interpreting have tended to move away from discussions towards examining expectations of how interpreters react in specific situations. The research of Kredens (2016), which sought responses from legal professionals and interpreters to vignettes representing possible ethical dilemmas in legal interpreting is a prime example. Here, legal professionals tended to talk about ethical decisions in light of legal procedure, while interpreters tended to concentrate on the practicalities of certain solutions and the need for professional distance (ibid, pp. 70-72).

This foreshadowed the later work of Machado and Downie (2024) on how technological changes during and after the pandemic in Brazil changed how legal interpreters view their own role and how legal professionals viewed interpreters. Once again, a mild Eraslan effect is evident. While both legal professionals and interpreters viewed the abstract role of interpreters in terms of the need for defendants from other countries to have a linguistic presence in the legal process (ibid, p. 206), they differed in what this meant. Legal professionals tended to favour the interpreter doing the best they could to avoid interrupting the legal process in any way (ibid, p. 208), while interpreters viewed the position of the interpreter as a mandate for greater professionalisation and societal change (ibid, p. 211).

Differences in role expectations and subsequent views of quality are also found in the survey research of Martínez-Goméz (2015: 183) on the expectations of non-professional interpreters in prisons, which concluded that the role of these interpreters was influenced by local institutional and societal factors. Prime among these was the conflict in expectations between fellow prisoners, who expected interpreters to function as cultural brokers, and prison staff, who expected prisoners to advocate on their behalf (ibid, p. 184).

### 3.2 Product-based research in interpreting quality

If surveys of interpreting users do not seem to provide a route towards a universal definition of quality, it would seem eminently sensible to perform some kind of product-based analysis. Such research is the closest methodologically to the use of AEMs.

### 3.2.1 Monologic interpreting: error counts and variable manipulation

The work of Henri Barik set the groundwork for much later discussion of product-oriented evaluations of interpreting quality. His research, carried out in the 1960s and first published in 1971, provided a comprehensive coding scheme for various sorts of differences between the source language version and the output of the interpreter (Barik 1971) in monologic interpreting. This work, alongside the work of David Gerver (1974) on the relationship between sound quality and interpreting, created a tradition for product-oriented studies of interpreting quality in monologic interpreting. It must be said, before discussing some selected examples of such research, that most research on the interpreted product is not concerned with evaluating or measuring quality. There is a long tradition of research using linguistic tools to examine both monologic and dialogic interpreted output (e.g. Hatim and Mason 1990; Roy 1999; Khachula *et al.* 2021). These studies are descriptive in nature, seeking to understand patterns in interpreted output and their possible causes, not to evaluate them.

Returning to evaluations, on the basis of the outputs of 65 final year interpreting students, Franţescu (2020) constructed a classification of interpreting errors, that covered problems with lexis, morphology, pronunciation, syntax, and reformulation. In this case, omissions were seen to be an almost universal error form, with their presence assumed to be caused by the inexperience of the students (ibid, p. 93). Around the same time as this study was released, the importance and causes of omissions were under debate. For example, Eraslan (2020) found that omissions could be caused by speakers with non-native accents. It seems that classifying errors as part of evaluation and explaining the sources and importance of those errors are two entirely different tasks.

Indeed, the difference in purported explanations for omissions is an example of a rather widespread problem with product-based evaluations. The same behaviour that may be classified as an error by one scholar can be read as context-driven decision-making by another. For example, in a study of monologic church interpreting, Karlik remarked that the interpreters routinely added cohesive markers to aid comprehension (2010: 172–173). Yet similar additions were seen as errors in the research of Makha and Phafoli in the monologic church interpreting they observed (2019: 156). It appears that counting or classifying errors without an account of the context in which they appear should now be read as a methodological weakness. Indeed, the importance of context has been at the heart of research on dialogic interpreting.

*3.2.2 Dialogic interpreting: from back translation to holistic assessments*

Quality is at the heart of the discussion of the relationship between interpreting quality and different remote interpreting setups in the AVIDICUS project by the team led by Sabine Braun (Braun and Taylor 2012; Braun 2013; Braun 2016; Braun 2017). In this work, while there are measures of intertextual linguistic correspondence (Braun and Taylor 2012: 107), these were placed within an analysis of wider sociolinguistic and paralinguistic issues such as changes in turn-taking, fatigue, and the interpreter's ability to see what was going on (ibid, p. 110-111).

Such a multi-method, holistic evaluation frames methods and analysis within the ways in which interpreting is or will be used in authentic settings. Indeed, from the outset, the work by Braun and colleagues assumed that there was a need for methods of evaluation to be coherent with the end use of the technology. Assessing the reliability of video-mediated remote interpreting in specific legal settings was mentioned as one of the three core aims of the project (Braun and Taylor 2012: 33).

Outside of Interpreting Studies, the measurement of the quality of interpreting provided by automated speech translation (AST) apps in dialogic settings has followed a strikingly similar course to the AVIDICUS project. In one case, while patients reported a much higher rate of satisfaction with the apps used than was reported from medical professionals (Hudelson and Chappuis 2024: 1098), these results came with important caveats. Results for European languages were much better than for non-European languages (ibid, p. 1099). More importantly, the outcomes of patient-doctor consultations using AST systems were found to be best when speakers used fluent, plain language, and always spoke in full sentences (ibid, pp. 1099-1100). The effective use of AST therefore seems to require behavioural and linguistic changes from those involved, changes that may not be required when using human interpreters. Hence why the medical professionals involved in the study felt that AST was only suitable for encounters that involved the exchange of factual information with few emotional or cultural overtones (ibid, p. 1099).

The aforementioned study therefore seems to have reached conclusions that would be at home in any discussion of interpreting within Interpreting Studies. Most notable was the finding that "MT has several disadvantages, including difficulty detecting contextual clues and translating non-standard language, cultural expressions and disfluency" (ibid, p. 1100). What we have therefore is a clear distinction drawn between semantic accuracy, as has been tested using automated metrics, and fitness for purpose, which is a contextual variable.

Even when commercially available AST systems were given the advantage of an experimental setup that allowed them to perform at their best, with low background noise and clearly separated sentences, they still failed to meet the level of human performance (Lee *et al.* 2023). The difference here was that the evaluation did not just cover semantic accuracy but looked at the importance of any changes within the specific clinical situation envisaged by the researchers (ibid, p. 2334). Once again, the weaknesses found in AST systems was due to contextual issues, specifically, the inability to adjust for the needs of two-way communication or to accept hesitations and disfluencies as a normal part of medical communication (ibid, p. 2337).

As similar results were found using different methods (e.g. Birkenbeuel *et al.* 2021), the findings of these papers on medical uses of ASTs are not their main contribution. In the context of any discussion of AEMs for evaluating interpreting, what these studies represent is a clear argument in favour of context-sensitive evaluations. In short, their findings include explicit statements in favour of exactly the same fitness-for-purpose approach to the evaluation of interpreting that could be arrived at from a close reading of existing interpreting theory. Despite showing little awareness of Interpreting Studies literature, these evaluations of AST in medical settings approach the subject from a remarkably similar angle to the experimental work of the AVIDICUS group.

## 4. Towards Situated Quality

Taking into account the existing literature, it would seem that there is a clear delineation between research that seeks to discover decontextualised quality criteria and research that seeks to understand how useful a given interpreting system is in a specific context. While AEMs are the paradigm case of the former, the AVIDICUS project and recent evaluations of AST in medical settings are representatives of the latter.

Whenever research leans towards decontextualised notions of quality, the results may not be applicable to any specific situation. That would seem to be a truism but, in the light of the growth of interpreting AEMs, it is an important statement to make. Indeed, Papi et al (2024) have shown that the predominance of AEMs that evaluate AST systems on the basis of pre-segmented data leads to a situation where the relationship between laboratory performance and real-life tests is unclear.

Thus, what is needed for quality assessments that provide anything more than a generic, and possibly contextually irrelevant, picture is a move towards what Buzungu and Hansen (2020: 60) call a "situatedness orientation". In other words, any ideas about quality must be localised and made relevant and appropriate to the specific situation in which quality is being measured. Quite simply, any attempt to measure interpreting quality must put the context in which the interpreting takes place at its heart. This moves far beyond asking humans to rate the quality of a specific stretch of text or using a specific metric and comparing that to automated ratings. Instead, the first and most powerful measurement of quality is how and to what extent the interpreting achieved its intended purpose(s) in the specific situation in which it was used.

This is more than a return to functionalism in Interpreting Studies (cf. Pöchhacker 1995). This "situatedness" orientation means shifting the emphasis away from general or even settings-based accounts of interpreting quality towards research that is carried out "on, for, and with" (Turner and Harrington 2000) the people involved in the interpreting situation. In short, research on quality must move away from decontextualised accounts towards evaluations of quality in specific contexts, using methods that make sense to and reflect the priorities of those involved. This is in complete agreement with Pöchhacker's call to place "agency, embodiment and immediacy" (2024: 15) at the heart of definitions of interpreting, especially since Pöchhacker argues that embodiment implies the situatedness of interpreting (ibid).

What part can interpreting AEMs play in this shift towards situated interpreting quality, especially given that such a shift would seem to run counter to the perceived need for fast quality assessments? Interpreting AEMs, if used in partnership with assessments that account for local quality factors and the interests of those involved, could provide an initial baseline impression of the kinds of challenges the interpreter faced or of the strictly mathematical relationship between the interpreter's production and the source texts they saw or heard.

In the light of the social nature of interpreting quality, it is logical to conclude that interpreting AEMs are not quality measurements in any sense that would accord with extant literature. Nor are they reliable measures of whether a given interpreting solution, delivered by human or machine or some combination of both, is offering an adequate level of service. Like the MT quality measurements on which some AEMs are based, interpreting AEMs can be used to compare different automated interpreting solutions and may be part of a broader assessment of quality. They are, however, tools with clear and specific limitations, most notably their inability to adjust to the needs of specific interpreting situations.

One further purpose for AEMs, which does not seem to have yet been explored, would be as a tool for hypothesis testing. The recent discussions of the commonalities among interpreting found in Comparative Interpreting Studies (Gile and Napier 2020; Buzungu and Hansen 2020; Downie 2021; Downie 2023) all use existing literature as the basis of their claims. AEMs, especially those which are tuned to evaluate measure aspects of quality, such as prosody, terminological accuracy, speed, and completeness, could be used to analyse large datasets to see if these claims actually match the data. In this sense, AEMs would be used to examine patterns in interpreting output and the relationship between source and target texts in interpreting in different contexts. By doing so, they may well contribute to discussions of quality that do offer the possibility of finding common, contextually relevant and measurable quality criteria that apply across all interpreting.

## 5. Conclusion and future directions

Examining the use of interpreting AEMs and comparing them with discussions of quality in Interpreting Studies research leads to the conclusion that interpreting AEMs are not viable measures of the quality of any interpreting provision when used on their own. Across all attempts to measure or even categorise quality in Interpreting Studies, the contexts in which interpreting takes place have become fundamental to the final analysis. What makes good interpreting in one situation may not suffice in another. This is now a basic theoretical and methodological assumption that has been justified in numerous ways, across many different studies. Quality assessments therefore must include an awareness of the primacy of the context of interpreting in their construction and analysis. This does not spell the end for interpreting AEMs but means that they must be used in combination with other methods or as a means of validating existing hypotheses in large-scale datasets. The quest for a repeatable, quick, and automated method of evaluation fails simply because interpreted situations are neither repeatable, automated nor amenable to anything but careful, considered analysis. In addition, the extant literature suggests that it should be a requirement for research using AEMs as a primary method to include a clear caveat to the effect that they should not be read as unmediated measures of the quality of the interpreting produced.